\documentclass[10pt,twocolumn,letterpaper]{article}

\usepackage{template/iccv}
\usepackage{times}
\usepackage{epsfig}
\usepackage{graphicx}
\usepackage{amsmath}
\usepackage{amssymb}

\usepackage{multirow}
\usepackage{tikz}

\usepackage{url}
\usepackage{ctable}
\usepackage{booktabs}
\usepackage{tabularx}
\usetikzlibrary{fit}
\usepackage{makecell}
\usepackage{colortbl}
\usepackage{pifont}

\usepackage{comment}
\usepackage{enumitem}
\usepackage[normalem]{ulem}
\usepackage{array}
\usepackage{textcomp}
\usepackage{mathtools}
\usepackage[flushmargin]{footmisc}
\setlist[itemize]{align=parleft,left=0pt,topsep=1mm,itemsep=0mm,parsep=1mm}
\usepackage{float}

\usepackage[pagebackref,breaklinks,colorlinks,citecolor=cyan,linkcolor=cyan,bookmarks=false]{hyperref}

\newcommand{\calD}{{\mathcal{D}}}

\newcommand{\calL}{{\mathcal{L}}}

\newcommand{\be}{\begin{eqnarray}}
\newcommand{\ee}{\end{eqnarray}}
\newcommand{\bee}{\begin{eqnarray*}}
\newcommand{\eee}{\end{eqnarray*}}

\newcommand{\matrixb}{\left[ \begin{array}}
\newcommand{\matrixe}{\end{array} \right]}

\newcommand{\app}{\raise.17ex\hbox{$\scriptstyle\sim$}}
\input{template/def_custom.tex}

\newcommand{\newpara}[1]{\vspace{6pt}\noindent\textbf{#1}}
\newcommand*{\myalign}[2]{\multicolumn{1}{#1}{#2}}
\definecolor{light_orange}{RGB}{252,219,191}
\definecolor{dark_orange}{RGB}{237,139,55}

\newcounter{nodecount}
\newcommand\tabnode[1]{\addtocounter{nodecount}{1} \tikz \node  (\arabic{nodecount}) {#1};}

\tikzstyle{every picture}+=[remember picture,baseline]
\tikzstyle{every node}+=[anchor=base,minimum width=0.4cm,align=center,text depth=.25ex,outer sep=1.5pt]
\tikzstyle{every path}+=[thick, rounded corners]

\iccvfinalcopy 

\def\iccvPaperID{9621} 

\ificcvfinal\fi

\begin{document}

\title{Sound Source Localization is All about Cross-Modal Alignment}

\author{Arda Senocak$^{1}$$^*$ \hspace{4mm}
Hyeonggon Ryu$^{1}$$^*$ \hspace{2mm}
Junsik Kim$^{2}$$^*$ \hspace{2mm} 
Tae-Hyun Oh$^{3,4}$ \hspace{2mm} \\
Hanspeter Pfister$^{2}$ \hspace{2mm}
Joon Son Chung$^{1}$ \vspace{5pt} \\
%
$^1$ Korea Advanced Institute of Science and Technology 
\hspace{5pt} $^2$ Harvard University
\\ $^3$Dept.~of Electrical Engineering and Grad.~School of Artificial Intelligence, POSTECH
\\ $^4$Institute for Convergence Research and Education in Advanced Technology, Yonsei University
%
}

\maketitle
\ificcvfinal\thispagestyle{empty}\fi

{\let\thefootnote\relax\footnote{$^*$These authors contributed equally to this work.}}
\thispagestyle{empty}

\begin{abstract}
Humans can easily perceive the direction of sound sources in a visual scene, termed sound source localization. 
Recent studies on learning-based sound source localization have mainly explored the problem from a localization perspective.
However, prior arts and existing benchmarks do not account for a more important aspect of the problem, cross-modal semantic understanding, which is essential for genuine sound source localization. 
Cross-modal semantic understanding is important in understanding semantically mismatched audio-visual events, \eg, silent objects, or off-screen sounds.
To account for this, we propose a cross-modal alignment task as a joint task with sound source localization to better learn the interaction between audio and visual modalities. 
Thereby, we achieve high localization performance with strong cross-modal semantic understanding. 
Our method outperforms the state-of-the-art approaches in both sound source localization and cross-modal retrieval.
Our work suggests that jointly tackling both tasks is necessary to conquer genuine sound source localization.
\end{abstract}
\vspace{-10mm}
\section{Introduction}\label{sec:intro}
Humans can easily perceive where the sound comes from in a scene.
We naturally attend to the sounding direction and associate incoming audio-visual signals to understand the event.
To achieve human-level audio-visual perception, sound source localization in  visual scenes has been extensively studied~\cite{senocak2018learning, senocak2019learning, arandjelovic2018objects, qian2020multiple, chen2021localizing, lin2021unsupervised, hu2020discriminative, li2021space, senocakLessMore, song2022sspl, senocakHardPos, ssslTransformation, ezvsl, slavc, htf}.
Motivated by that humans learn from natural audio-visual correspondences without explicit supervision, most of the studies have been developed on a fundamental assumption that audio and visual signals are temporally correlated. 
With the assumption, losses of the sound source localization task are modeled by audio-visual correspondence as a self-supervision signal and are implemented by contrasting audio-visual pairs, \ie, contrastive learning.

\begin{figure}[t]
\centering
        \includegraphics[width=1.0\linewidth]{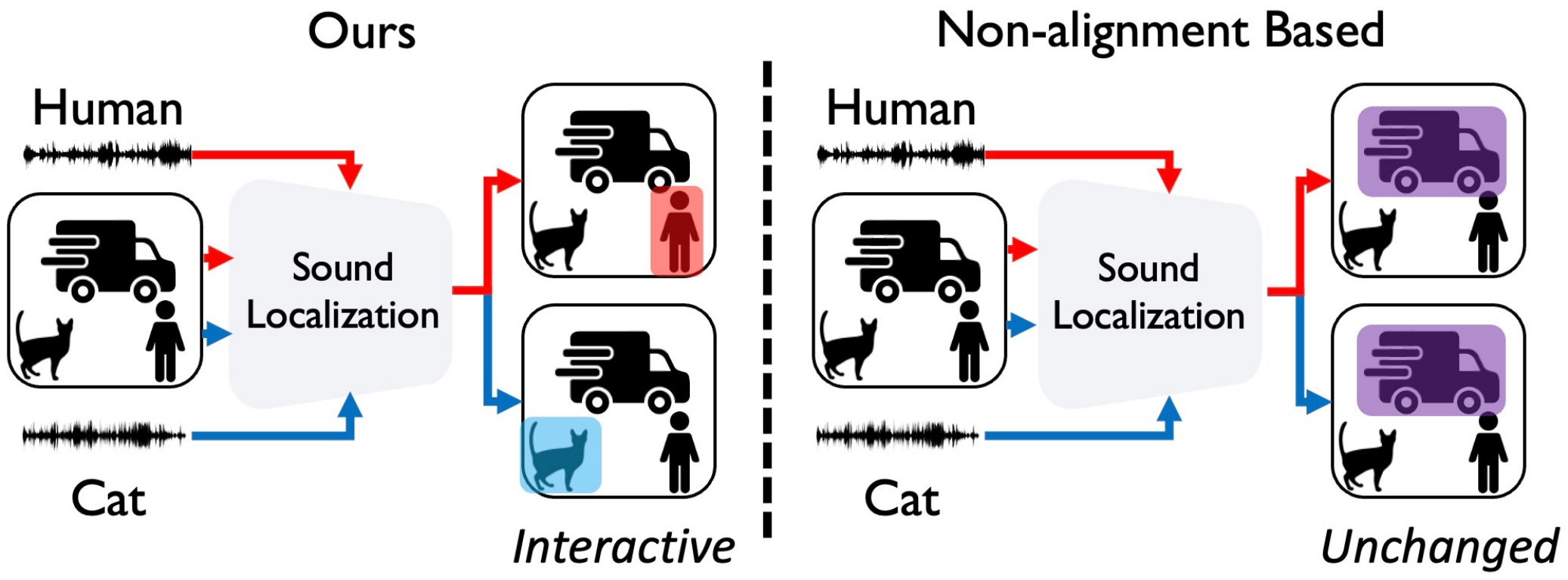}
\caption{\textbf{A conceptual difference between prior approaches and our alignment-based sound source localization.} 
}
\label{fig:second_teaser}
\vspace{-4mm}
\end{figure}

While these approaches appear to be unsupervised methods, they strongly rely on partial supervision information; \eg, using supervisedly pretrained vision networks~\cite{senocak2018learning, senocak2019learning, qian2020multiple, senocakLessMore, song2022sspl, htf} and visual objectness estimators for post-processing~\cite{ezvsl, slavc}.
Without leveraging such strong initial representations, the performance is degraded. Thus, the previous methods are not purely self-supervised approaches.
Even further, there are recent studies~\cite{oya2020we, ezvsl, slavc} that point out visual objectness bias in existing sound source localization benchmarks and exploit the objectness prior to improve the localization accuracy.
They show that, even without interaction between visual and audio signals, a model may achieve strong accuracy in localization by only referring visual signals alone, which is not the true intention of the sound source localization task. In short, the current evaluation and setting of the sound source localization do not capture the true sound source localization performance.

In this work, we first sort out evaluating sound source localization methods by introducing a cross-modal retrieval task as an auxiliary evaluation task.
By this task, we can measure whether the learned representation have the capability to accurately interact between audio and visual modalities; \ie, more fine-grained audio-visual correspondence which is essential for genuine sound source localization.
This aspect has been missed in existing sound source localization benchmarks.
Indeed, our experiments show that higher sound localization performance does not guarantee higher cross-modal retrieval performance. 

Second, given this additional criterion, we revisit the importance of semantic understanding shared across audio and visual modalities in both sound source localization and cross-modal retrieval.
In the previous methods~\cite{senocak2018learning,senocak2019learning,song2022sspl,qian2020multiple}, the cross-modal semantic alignment is induced by instance-level cross-modal contrastive learning, \ie, cross-modal instance discrimination between visual and audio features.
However, they are aided by labels or supervisedly pretrained encoder~\footnote{Typically, an image encoder is pretrained on ImageNet~\cite{deng2009imagenet} and an audio encoder is pretrained on AudioSet~\cite{audioset} in supervised ways.} for easing challenging cross-modal feature alignment.
Instead, 
our method learns from scratch supporting the lack of guidance by incorporating multiple 
positive samples into cross-modal contrastive learning. 
Specifically, we construct a positive set for each modality using both multi-view~\cite{chen2020simple} and conceptually similar samples~\cite{dwibedi2021little}. Thereby, we enhance feature alignment and achieve high localization performance and strong cross-modal semantic understanding.

We evaluate our method on the VGG-SS and SoundNet-Flickr benchmarks for sound source localization and cross-modal retrieval. 
As aforementioned, the sound source localization task is closely related to the cross-modal retrieval task, but our experiments show that existing works have a weak performance correlation between them. 
This implies that we need to evaluate both tasks for evaluating the genuine sound source localization. The proposed method performs favorably against the recent state-of-the-art approaches in both tasks.

We summarize the contributions of our work as follows:
\begin{itemize}
    \item We analyze that sound source localization benchmarks are not capable of evaluating cross-modal semantic understanding, thereby sound source localization methods may perform poorly in cross-modal retrieval tasks.
    \item We propose semantic alignment to improve cross-modal semantic understanding of sound source localization models.
    \item We expand semantic alignment with multi-views and conceptually similar samples which leads to state-of-the-art performance on both sound source localization and cross-modal retrieval. 
\end{itemize}


\vspace{-4mm}
\section{Related work}\label{sec:RW}
\vspace{-4mm}
\newpara{Sound source localization.} Sound source localization in visual scenes has been investigated by exploiting correspondences between audio and visual modalities.
The most widely used approach for sound source localization is cross-modal attention~\cite{senocak2018learning, senocak2019learning, tian2018audio} with contrastive loss~\cite{chopra2005learning, hoffer2015deep, infoNCE}. Later, the attention-based method is improved 
by intra-frame hard sample mining~\cite{chen2021localizing}, iterative contrastive learning with pseudo labels~\cite{lin2021unsupervised}, 
feature regularization~\cite{ssslTransformation}, positive mining~\cite{senocakHardPos}, negative free learning~\cite{song2022sspl} with stop-gradient operation~\cite{chen2021exploring}, or momentum encoders~\cite{slavc}.

Some sound localization approaches exploit additional semantic labels~\cite{qian2020multiple, li2021space, senocakLessMore} or object prior~\cite{ezvsl, xuan2022proposal}. Semantic labels are used to pretrain audio and vision encoders with classification loss~\cite{li2021space, senocakLessMore} or refine audio-visual feature alignment~\cite{qian2020multiple}. A more explicit way to refine localization output is to use object prior. EZVSL~\cite{ezvsl} proposes post-processing to combine attention based localization output with a pretrained visual feature activation map. Similarly, Xuan \etal~\cite{xuan2022proposal} propose to combine off-the-shelf object proposals with attention based sound localization results.
However, postprocessing by object prior may generate a false positive output as it is solely based on vision without audio-visual interaction.

In addition to the localization, there has been an attempt to localize sounding objects and recover the separated sounds simultaneously, also known as the cocktail party problem~\cite{haykin2005cocktail, mcdermott2009cocktail}. The separation of sound mixture is achieved by predicting masks of spectrogram guided by visual features~\cite{ephrat2018looking, afouras2018conversation, zhao2019sound, gao2019coSep, xu2019minusPlus, gan2020gestureSep, afouras2020AVObjects, zhou2020sepStereo, gao2021visualVoice, tzinis2021audioScope, tian2021cyclic}. Furthermore, a number of recent papers are presented on audio-visual navigation for a given sound source~\cite{chen2020soundspaces, gan2020look}.

\begin{figure*}[tp]
    \centering
    \includegraphics[width=\linewidth]{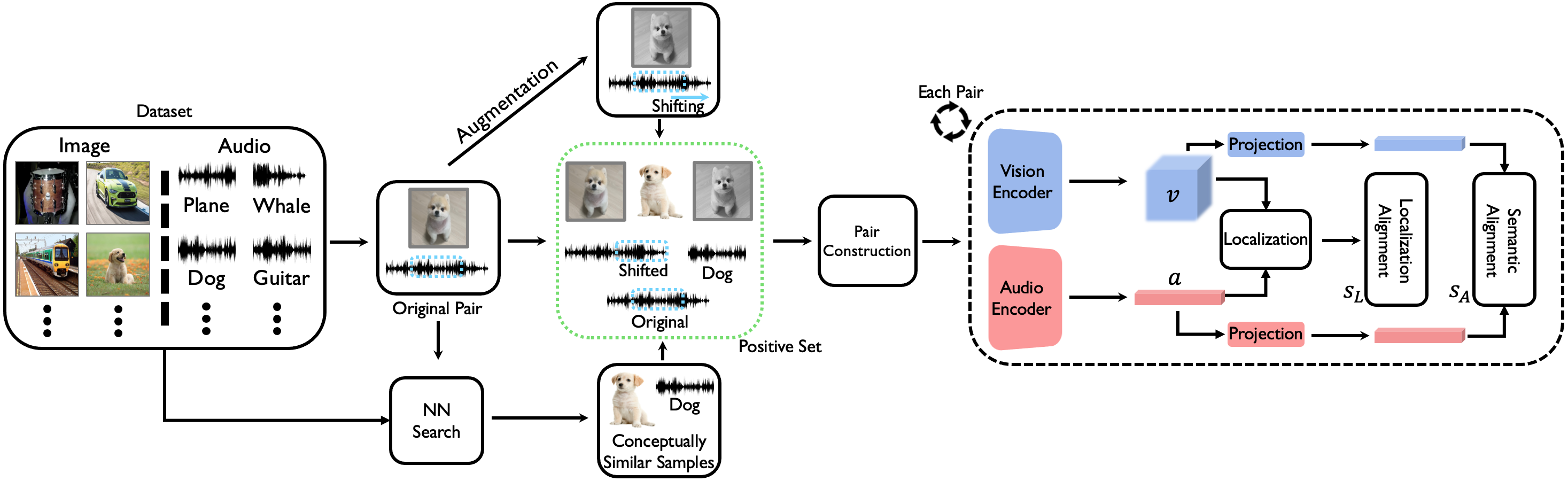}
    \caption{{\bf Our sound source localization framework.} Our model construct multiple positive pairs with augmentation and Nearest Neighbor Search (Conceptually Similar Samples). By using these newly constructed 9 pairs, our model employs spatial localization, $s_L$, and semantic feature alignment, $s_A$, for each pair to learn a better sound source localization ability.}
    \label{fig:pipeline}
\end{figure*}

\newpara{Self-supervised representation learning.} In a broader categorization, sound source localization belongs to self-supervised multimodal learning. Our work is also relevant to self-supervised audio-visual representation learning, and other multimodal learning studies.

Contrastive learning aims to learn robust representations from large-scale raw data without annotations. Recent representation learning approaches~\cite{wu2018unsupervised, chen2020simple, he2020momentum, chen2020improved} use instance discrimination by contrastive learning~\cite{chopra2005learning, hoffer2015deep, infoNCE} as a pretext task with notable advancements in visual recognition tasks. Recently, positive mining by nearest-neighbor search are used to learn representations of images~\cite{dwibedi2021little, elbanani2022languageguided, xu2022seed}, videos~\cite{han2020coclr}, neural recordings~\cite{azabou2021mine},
and text-image~\cite{li2021supervision}. In this work, we expand the previous works by incorporating both multi-views and conceptually similar samples into audio-visual modalities for cross-modal feature alignment.

A series of audio-visual representation learning studies have shown that audio and visual contents in a video are correlated, therefore a visual representation can be learned by sound prediction~\cite{owens2016ambient} or audio representation can be distilled from visual representation~\cite{aytar2016soundnet,sung2023sound}. Later, a variety of joint audio-visual representation learning methods are proposed with an assumption that there is a semantic~\cite{arandjelovic2017look, hu2019deep, morgado2021audio, morgado2021robust} or temporal~\cite{chung2017out,Owens2018AudioVisualSA, korbar2018cooperative,chung2020seeing} correspondence between them. However, simply learning sound source localization by audio-visual correspondence with instance discrimination ignores the semantic similarity of audio-visual contents among samples, introducing false negatives or positives. 
In order to mitigate this issue, clustering~\cite{hu2019deep}, sampling~\cite{morgado2021audio}, weighting~\cite{morgado2021robust}, and hard mining~\cite{korbar2018cooperative} are proposed.
Similarly, in this work, we go beyond instance discrimination by using multiple positive samples to enforce semantic understanding across modalities.

\section{Method}\label{sec:MTD}
\subsection{Preliminaries} \label{ssec:pre}

\newpara{Contrastive learning} learns representation by containing positive and negative pairs.
Given an encoded query sample $q$ and its encoded positive pair $k^+$ and negative pairs $k$, the loss can be defined as:
\begin{equation}
\label{eq:contrastive}
\begin{aligned}
        \calL=\mathrm{-log}\frac{\mathrm{exp}(q\cdot k^+/\tau)}{\sum_{i}\mathrm{exp}(q\cdot k_i/\tau)}
\end{aligned}
\end{equation}
where $\tau$ is the temperature parameter.

\newpara{Cross-modal contrastive learning}
extends contrastive learning across multiple modalities.
In sound source localization, audio-visual correspondence is used to define positive and negative cross-modal pairs. 
With an audio-visual dataset $\calD=\{(v_i,a_i):i=1,...,N\}$ and its encoded features $\mathbf{v}_i=f_v(v_i)$ and $ \mathbf{a}_i=f_a(a_i)$, cross-modal contrastive learning loss is defined as: 
\begin{equation}
\label{eq:av_contrastive}
\begin{aligned}
        \calL_i=\mathrm{-log}\frac{\mathrm{exp}(s(\mathbf{v}_i, \mathbf{a}_i)/\tau)}{\sum_{j}\mathrm{exp}(s(\mathbf{v}_i, \mathbf{a}_j)/\tau)}
\end{aligned}
\end{equation}
where $s$ is a cross-modal similarity function. The cross-modal contrastive loss~\Eref{eq:av_contrastive} can be extended to symmetric form~\cite{radford2021learning} as used in a few previous works~\cite{ezvsl,slavc}.

\subsection{Cross-Modal Feature Alignment} \label{ssec:feat_align}
We consider both spatial localization and semantic feature alignment for sound source localization.
To this end, we use two different similarity functions $s_L$ and $s_A$ for contrastive learning~(\Eref{eq:av_contrastive}), $s_L$ for localization and $s_A$ for cross-modal feature alignment.

Recent studies rely on audio-visual spatial correspondence maps to learn sound source localization by contrasting them.
Given a spatial visual feature $\mathbf{v} \in \mathbb{R}^{c\times h\times w}$
and audio feature $\mathbf{a \in \mathbb{R}^{c}}$, audio-visual similarity with a correspondence map can be calculated as follows:
\begin{equation}
    \label{eq:sim_localize}
    \begin{aligned}
            s_{L}(\mathbf{v}, \mathbf{a}) = \sum_{xy \in M} \frac{1}{|M|} \frac{\mathbf{v}^{xy} \cdot \mathbf{a}}{\|\mathbf{v}^{xy}\| \| \mathbf{a}\|}
    \end{aligned}
\end{equation}
where $\mathbf{v}^{xy}$ is a feature vector at location $(x,y)$, and $M$ is an optional binary mask when an annotation or pseudo-mask~\cite{chen2021localizing, ssslTransformation} is available.
Since we assume no supervision for sound source localization, we do not use any mask, therefore, $M=\mathbf{1}$. 

The contrastive loss with localization similarity $s_L$ enforces location dependent alignment giving sparse but strong audio-visual correspondence which enables to perform localization. 
However, our empirical studies on cross-modal retrieval indicate that strong localization performance does not guarantee semantic understanding.
To overcome the low semantic understanding in recent studies, we propose to add instance-level contrastive loss.
Instance-level contrasting encapsulates the whole context in a scene, enforcing better audio-visual semantic alignment. 
However, instance-level contrasting may smooth out spatial discriminativeness learned by \Eref{eq:sim_localize}. Inspired by SimCLR~\cite{chen2020simple}, we adopt a projection layer to align audio-visual semantics in a projection space. 
The projection layer separates the latent space of localization and semantic alignment, thereby preventing the alignment loss smoothing out the spatial discriminativeness.
The similarity function for cross-modal feature alignment is defined as follows:
\begin{equation}
    \label{eq:sim_align}
    \begin{aligned}
            s_{A}(\mathbf{v}, \mathbf{a}) = \frac{p_v(\mathsf{avg}\mathbf{(v)}) \cdot p_a(\mathbf{a})}{\|p_v(\mathsf{avg}(\mathbf{v}))\| \| p_a{\mathbf{a}}\|}
    \end{aligned}
\end{equation}
where $\mathsf{avg}(\cdot)$ is spatial average pooling, $p_v$ is a projection layer for visual features, and $p_a$ is a projection layer for audio features.

\subsection{Expanding with Multiple Positive Samples} \label{ssec:multiple_positives}
Typically, contrastive learning contrasts between one positive pair and multiple negative pairs as shown in~\Eref{eq:contrastive}.
In audio-visual learning, by an audio-visual correspondence assumption, an audio-image pair from the same clip is used as a positive pair while negative pairs are sampled from different clips. 
However, single-instance discrimination may not be sufficient to achieve strong cross-modal alignment.
In this section, we expand contrastive learning beyond single instance discrimination by positive set construction and pairing them. 
To construct a positive set, we incorporate both hand-crafted positive and conceptual positive samples for each modality.
Later, we adjust the contrastive learning to incorporate multiple positive pairs to enforce cross-modal alignment.

\newpara{Obtaining hand-crafted positive samples.}
Using randomly augmented samples as positive multi-view pairs are widely adopted in self-supervised representation learning, \ie, instance discrimination. 
Similarly, we extend a single anchor audio-image pair to multiple positive pairs by applying simple augmentations on image and audio samples separately. While we utilize common image transformations on images, we apply temporal shifting to audios. It is worth noting that sound source localization task learns from the underlying semantic consistency rather than subtle time differences as in videos. Thus, a slight shift in the audio may not alter contextual information significantly. As a result of hand-crafted multi-view positive pair generation, we obtain additional $\mathbf{v}^{aug}$ and $\mathbf{a}^{aug}$ samples.

\newpara{Obtaining conceptual positive samples.} 
Apart from manually created augmented views, we additionally expand our positive set with conceptually similar samples. 
The sampling strategy with nearest neighbor search can be performed in a various way, such as on-the-fly sampling~\cite{dwibedi2021little, ryu2023hindi,xu2022seed, li2021supervision}, sampling by pretrained encoders~\cite{senocakHardPos}, or guided sampling~\cite{han2020coclr, elbanani2022languageguided} using another modality. For selecting our conceptually similar samples, we utilize pretrained encoders. Note that pretrained encoders trained either with supervised or self-supervised learning are effective in positive sample mining as shown in the experiment section. By employing readily available image and audio encoders, we use the $k$-nearest neighborhood search to sample semantically similar samples in both modalities. In particular, given a pair of image and audio, we compute cosine similarity with all other samples and choose the top-$k$ most similar samples among the training set for each modality. From a set of $k$ samples, we randomly select one sample to obtain conceptually similar samples for each modality, $\mathbf{v}^{conc.}$ and $\mathbf{a}^{conc.}$. By utilizing the conceptually similar samples as positive samples, our model expands semantic understanding. 

\newpara{Pair Construction.} Once we obtain the conceptual and hand-crafted positive samples for each modality, we proceed to create 9 distinct audio-visual pairs by pairing $\mathbf{V}=\{\mathbf{v}, \mathbf{v}^{aug}, \mathbf{v}^{conc}\}$ and $\mathbf{A}=\{\mathbf{a}, \mathbf{a}^{aug}, \mathbf{a}^{conc}\}$. This is done to ensure semantic alignment and consistency between them through contrastive learning.
The negative pairs are randomly paired from the remaining samples in a training set. It is worth noting that some of these pairs are a combination of hand-crafted and conceptually similar samples, which further enhances the feature alignment of our model during training.

\subsection{Training} \label{ssec:feat_align}

Our loss formulation incorporates both localization and instance-level similarity functions with multiple positive pairs constructed by augmentation and conceptually similar sample search. 
The final loss term is defined as follows:
\begin{equation}
\label{eq:loss}
\begin{aligned}
        \calL_i=-
        \sum_{\mathbf{v}_i \in \mathbf{V}}
        \sum_{\mathbf{a}_i \in \mathbf{A}}
        \left[
        \mathrm{log}\frac{\mathrm{exp}(s_L(\mathbf{v}_i, \mathbf{a}_i)/\tau)}{\sum_{j}\mathrm{exp}(s_L(\mathbf{v}_i, \mathbf{a}_j)/\tau)} \right.
        \\+ \left.
        \mathrm{log}\frac{\mathrm{exp}(s_A(\mathbf{v}_i, \mathbf{a}_i)/\tau)}{\sum_{j}\mathrm{exp}(s_A(\mathbf{v}_i, \mathbf{a}_j)/\tau)}
        \right]
\end{aligned}
\end{equation}
where $\mathbf{V}$ and $\mathbf{A}$ indicate positive sample sets.

\section{Experiments}
Our proposed method for sound source localization is validated through experiments conducted on VGGSound~\cite{VGGSound} and SoundNet-Flickr~\cite{aytar2016soundnet}. First, we conduct a quantitative analysis to evaluate the accuracy of the localization, cross-modal retrieval, and the impact of various components of our model. Then, we visualize our sound source localization results across different categories of sounds.

\subsection{Experiment Setup}\label{ssec:exp_setup}
\newpara{Datasets.} Our method is trained using the VGGSound~\cite{VGGSound} and SoundNet-Flickr-144K~\cite{senocak2018learning,senocak2019learning}. VGGSound is an audio-visual dataset containing around \app 200K videos. SoundNet-Flickr-144K set is the subset of SoundNet-Flickr~\cite{aytar2016soundnet}. After training, we test the sound localization performance with VGG-SS~\cite{chen2021localizing} and SoundNet-Flickr-Test~\cite{senocak2018learning} datasets for the main experiments. These evaluation sets have bounding box annotations of sound sources for \app 5K and 250 samples, respectively. Moreover, we employ the AVSBench~\cite{zhou2022avs} and Extended VGGSound/SoundNet-Flickr~\cite{slavc} datasets for additional evaluations. AVSBench dataset provides binary segmentation maps that show the audio-visually correspondent pixels for roughly 5k five-second videos belonging to 23 categories. Lastly, the Extended VGGSound /SoundNet-Flickr dataset, proposed by~\cite{slavc}, is used to understand non-visible sound sources. 

\newpara{Implementation details.} We use two ResNet18 models for both audio and vision encoding. Unlike prior approaches, we do not fine-tune (or use a pretrained) a visual encoder from ImageNet pretrained weights.
Instead, we train both the audio and vision encoders from scratch. 
We preprocess images and audios following the previous works~\cite{chen2021localizing, senocakHardPos}.
To create multiple pairs, we utilize both NN search and generic augmentation approaches. For NN search, we experiment on two different setups to retrieve k conceptually similar samples: (1) For supervisedly pretrained encoder experiments, We employ ResNet and VGGSound models pretrained on ImageNet and VGGSound respectively, (2) For self-supervisedly pretrained encoder experiments, we utilize the CLIP~\cite{radford2021learning} Vision Encoder and Wav2CLIP~\cite{wu2022wav2clip} Audio Encoder.
We use $k$=1000 for the experiments. To perform image augmentations, we follow the augmentations used in SimCLR~\cite{chen2020simple}.
For audios, we randomly select time-window shifts in a time axis. 
The model is trained for 50 epochs with Adam Optimizer and a learning rate of 0.0001. $\tau$ is set to 0.07 in contrastive learning.

\begin{table}
    \centering
    \resizebox{1.0\linewidth}{!}{
    \begin{tabular}{lcccccc}
    \toprule
    &\multicolumn{1}{c}{}& \multicolumn{2}{c}{\textbf{VGG-SS}} & \multicolumn{2}{c}{\textbf{Flickr-SoundNet}} \\
    \textbf{Method} & \textbf{Pre. Vision} &  \textbf{cIoU $\uparrow$} & \textbf{AUC $\uparrow$} & \textbf{cIoU $\uparrow$} & \textbf{AUC $\uparrow$} \\ \midrule
    Attention~\cite{senocak2018learning}$_{\text{CVPR}18}$ & \ding{51}  &  18.50 & 30.20 & 66.00 & 55.80 \\
    CoarseToFine~\cite{qian2020multiple}$_{\text{ECCV}20}$  & \ding{51} 	 & 29.10 & 34.80 & - & - \\
    LCBM~\cite{senocakLessMore}$_{\text{WACV}22}$ & \ding{51}  	 & 32.20 & 36.60 & - & - \\
    LVS~\cite{chen2021localizing}$\dagger$$_{\text{CVPR}21}$ & \ding{55}   & 30.30 & 36.40 & 72.40 & 57.80 \\
    LVS~\cite{chen2021localizing}$_{\text{CVPR}21}$ & \ding{55}   & 34.40 & 38.20 & 71.90 & 58.20 \\
    HardPos~\cite{senocakHardPos}$_{\text{ICASSP}22}$ & \ding{55}   & 34.60 & 38.00 & 76.80 & 59.20 \\
    SSPL (w/o PCM)~\cite{song2022sspl}$_{\text{CVPR}22}$ & \ding{51}   & 27.00 & 34.80 & 73.90 & 60.20 \\
    SSPL (w/ PCM)~\cite{song2022sspl}$_{\text{CVPR}22}$ & \ding{51}   & 33.90 & 38.00 & 76.70 & 60.50 \\
    EZ-VSL (w/o OGL)~\cite{ezvsl}$_{\text{ECCV}22}$ & \ding{51}   & 35.96 & 38.20 & 78.31 & 61.74 \\
    SSL-TIE~\cite{ssslTransformation}$_{\text{ACM MM}22}$ & \ding{55}   & 38.63 & 39.65 & 79.50 & 61.20 \\
    SLAVC (w/o OGL)~\cite{slavc}$_{\text{NeurIPS}22}$& \ding{51}  & 37.79 & 39.40 & \textbf{83.60} & - \\
    \rowcolor{lightgray!25}
    \myalign{l}{\hspace{-0.6em}\tabnode{\textbf{Ours}}} & & & & & \\
    \rowcolor{lightgray!25}
    \myalign{l}{\;\;\;\footnotesize $\rotatebox[origin=c]{180}{$\Lsh$}$ NN Search w/ Supervised Pre. Encoders} & \ding{55} & \textbf{39.94} & \textbf{40.02} & \underline{79.60} & \textbf{63.44} \\
    \rowcolor{lightgray!25}
    \myalign{l}{\;\;\;\footnotesize $\rotatebox[origin=c]{180}{$\Lsh$}$ NN Search w/ Self-Supervised Pre. Encoders} & \ding{55} & \underline{39.20} & \underline{39.70} & 79.20 & \underline{63.00} \\
    \bottomrule
    \textit{with OGL:} &    &  &  &  &  \\
    EZ-VSL (w/ OGL)~\cite{ezvsl}$_{\text{ECCV}22}$ & \ding{51}   & 38.85 & 39.54 & \underline{83.94} & 63.60 \\
    SLAVC (w/ OGL)~\cite{slavc}$_{\text{NeurIPS}22}$& \ding{51}  & 39.80 & - & \textbf{86.00} & - \\
    \rowcolor{lightgray!25}
    \myalign{l}{\hspace{-0.6em}\tabnode{\textbf{Ours (w/ OGL)}}} & & & & & \\
    \rowcolor{lightgray!25}
    \myalign{l}{\;\;\;\footnotesize $\rotatebox[origin=c]{180}{$\Lsh$}$ NN Search w/ Supervised Pre. Encoders} & \ding{55} & \textbf{42.64} & \textbf{41.48} & 82.40 & \textbf{64.60} \\
    \rowcolor{lightgray!25}
    \myalign{l}{\;\;\;\footnotesize $\rotatebox[origin=c]{180}{$\Lsh$}$ NN Search w/ Self-Supervised Pre. Encoders} & \ding{55} & \underline{42.47} & \underline{41.42} & 82.80 & \underline{64.48} \\
    \bottomrule
    \textit{with Optical Flow:} &    &  &  &  &  \\
    HearTheFlow~\cite{htf}$_{\text{WACV}23}$& \ding{51}  & 39.40 & 40.00 & 84.80 & 64.00 \\
    \bottomrule
    \end{tabular}}
    
    {
    \caption{\textbf{Quantitative results on the VGG-SS and SoundNet-Flickr test sets}. All models are trained with 144K samples from VGG-Sound and tested on VGG-SS and SoundNet-Flickr. $\dagger$ is the result of the model released on the official project page. SLAVC~\cite{slavc} does not provide AUC scores.}\label{tab:quantitative}}
    \vspace{-2mm}
\end{table}

\subsection{Quantitative Results} \label{ssec:quan}
\newpara{Comparison with strong baselines.} In this section, we conduct a comparative analysis of our sound source localization method against existing approaches. We carry out our evaluations in two settings, following previous approaches. Firstly, we train our model on VGGSound-144K and evaluate it on VGG-SS and SoundNet-Flickr test sets. Secondly, we train our model on SoundNet-Flickr-144K and evaluate it on the SoundNet-Flickr test set. It is important to note that all the compared models are trained using the same amount of data. AVEL~\cite{tian2018audio}, AVObject~\cite{afouras2020AVObjects}, and LCBM~\cite{senocakLessMore} models rely on video input, and as such, they cannot be evaluated on the SoundNet-Flickr dataset, which contains static image and audio pairs. We present our results in~\Tref{tab:quantitative} and~\Tref{tab:quantitative_second}.

\begin{table}
    \centering
    \resizebox{1.0\linewidth}{!}{
    \begin{tabular}{lccc}
    \toprule
    \textbf{Method}     & \textbf{Pre. Vision}  & \textbf{cIoU $\uparrow$}    & \textbf{AUC $\uparrow$} \\ \midrule
    Attention\cite{senocak2018learning}$_{\text{CVPR}18}$ & \ding{51} & 66.00 & 55.80 \\
    DMC\cite{hu2019deep}$_{\text{CVPR}19}$ & \ding{51} & 67.10 & 56.80 \\
    LVS~\cite{chen2021localizing}$\dagger$$_{\text{CVPR}21}$  & \ding{55}   & 67.20 & 56.20 \\
    LVS~\cite{chen2021localizing}$_{\text{CVPR}21}$ & \ding{55}  & 69.90 & 57.30 \\
    HardPos~\cite{senocakHardPos}$_{\text{ICASSP}22}$ & \ding{55} & 75.20 & 59.70 \\
    SSPL (w/o PCM)~\cite{song2022sspl}$_{\text{CVPR}22}$ & \ding{51}   & 69.90 & 58.00 \\
    SSPL (w/ PCM)~\cite{song2022sspl}$_{\text{CVPR}22}$ & \ding{51}   & 75.90 & 61.00 \\
    EZ-VSL (w/o OGL)~\cite{ezvsl}$_{\text{ECCV}22}$ & \ding{51}   & 71.89 & 58.81 \\
    SSL-TIE~\cite{ssslTransformation}$_{\text{ACM MM}22}$ & \ding{55}   & 81.50 & 61.10 \\
    SLAVC (w/o OGL)~\cite{slavc}$_{\text{NeurIPS}22}$& \ding{51}  & - & - \\
    \rowcolor{lightgray!25}
    \myalign{l}{\hspace{-0.6em}\tabnode{\textbf{Ours}}} & & & \\
    \rowcolor{lightgray!25}
    \myalign{l}{\;\;\;\footnotesize $\rotatebox[origin=c]{180}{$\Lsh$}$ NN Search w/ Supervised Pre. Encoders} & \ding{55} & \textbf{85.20} & \underline{62.20} \\
    \rowcolor{lightgray!25}
    \myalign{l}{\;\;\;\footnotesize $\rotatebox[origin=c]{180}{$\Lsh$}$ NN Search w/ Self-Supervised Pre. Encoders} & \ding{55} & \underline{84.80} & \textbf{62.66} \\
    \bottomrule
    \textit{with OGL:} &    &  &   \\
    EZ-VSL (w/ OGL)~\cite{ezvsl}$_{\text{ECCV}22}$ & \ding{51}   & 83.13 & 63.06 \\
    SLAVC (w/ OGL)~\cite{slavc}$_{\text{NeurIPS}22}$ & \ding{51}  & - & - \\
    \rowcolor{lightgray!25}
    \myalign{l}{\hspace{-0.6em}\tabnode{\textbf{Ours (w/ OGL)}}} & & & \\
    \rowcolor{lightgray!25}
    \myalign{l}{\;\;\;\footnotesize $\rotatebox[origin=c]{180}{$\Lsh$}$ NN Search w/ Supervised Pre. Encoders} & \ding{55} & \underline{84.00} & \underline{64.16} \\
    \rowcolor{lightgray!25}
    \myalign{l}{\;\;\;\footnotesize $\rotatebox[origin=c]{180}{$\Lsh$}$ NN Search w/ Self-Supervised Pre. Encoders} & \ding{55} & \textbf{84.40} & \textbf{64.38} \\
    \bottomrule
    \textit{with Optical Flow:} &    &  &   \\
    HearTheFlow~\cite{htf}$_{\text{WACV}23}$ & \ding{51} & 86.50 & 63.90 \\
    \bottomrule
    \end{tabular}}
    {
    \caption{\textbf{Quantitative results on the SoundNet-Flickr test set.} All models are trained and tested on the SoundNet-Flickr 144K dataset. $\dagger$  is the result of the model from the official project page. SLAVC~\cite{slavc} does not provide results with SoundNet-Flickr 144K.}\label{tab:quantitative_second}}
    \vspace{-2mm}

\end{table}

Our proposed model achieves higher performance compared to prior approaches on both test sets. Specifically, it yields a +2.15$\%$ cIoU and +0.6$\%$ AUC improvement on VGGSS, as well as a +3.7$\%$ cIoU improvement on SoundNet-Flickr compared to the state-of-the-art methods that uses pretrained vision encoder. It is worth highlighting that unlike the majority of previous works, our proposed model does not utilize a vision encoder pretrained on ImageNet in a sound source localization backbone. This is because, as discussed in Mo \etal~\cite{slavc}, using supervisedly pretrained vision encoders makes the sound source localization problem a weakly supervised problem. However, it is worth noting that even without using a pretrained vision encoder, our method achieves state-of-the-art performance on both experiments that are presented in~\Tref{tab:quantitative} and~\Tref{tab:quantitative_second}. We demonstrate the performance of our model with the pretrained models learned through supervised learning (NN Search w/ Supervised Pre. Encoders) and with models that are pretrained through self-supervised learning (NN Search w/ Self-Supervised Pre. Encoders) in NN Search module. As the results indicate, using self-supervised pretrained encoders in NN Search performs on par with the supervised pretrained encoders in NN Search. This shows that our model does not depend on supervised pretrained encoders for the NN search module and can utilize any type of pretrained encoder feature for nearest neighbor search. Note that these pretrained encoders are not used in the backbone networks of the sound source localization module but only in the NN Search Module, as illustrated in~\Fref{fig:pipeline}.

\begin{table}
    \centering
    \footnotesize
    \scalebox{0.75}{
    \begin{tabular}{l|lccc}
    \toprule
    \textbf{Test Class} & \textbf{Method}   & \textbf{Pre. Vision}  & \textbf{cIoU $\uparrow$}  & \textbf{AUC $\uparrow$} \\ 
    \hline 
    \multirow{9}{*}{Heard 110}    
    & LVS~\cite{chen2021localizing}$_{\text{CVPR}21}$ & \ding{55} &28.90 & 36.20 \\
    & EZ-VSL(w/o OGL)~\cite{ezvsl}$_{\text{ECCV}22}$ & \ding{51} &31.86 & 36.19 \\
    & SLAVC(w/o OGL)~\cite{slavc}$_{\text{NeurIPS}22}$ & \ding{51} & 35.84 & - \\
    & \textbf{Ours} &  \ding{55} &\textbf{38.31} & \textbf{39.05}\\
    \cline{2-5}
    & \textit{with OGL:}&  &  &  \\
    & EZ-VSL(w/ OGL)~\cite{ezvsl}$_{\text{ECCV}22}$ & \ding{51} &37.25 & 38.97\\
    & SLAVC(w/o OGL)~\cite{slavc}$_{\text{NeurIPS}22}$ & \ding{51} & 38.22 & - \\
    & \textbf{Ours(w/ OGL)} & \ding{55} &\textbf{41.85} & \textbf{40.93} \\
    \cline{2-5}
    & \textit{with Optical Flow:}&  &  &  \\
    & HearTheFlow~\cite{htf}$_{\text{WACV}23}$& \ding{51} & 37.30 & 38.60 \\
    \hline \hline

    \multirow{9}{*}{Unheard 110} 
    & LVS~\cite{chen2021localizing}$_{\text{CVPR}21}$ & \ding{55} &26.30 & 34.70 \\
    & EZ-VSL(w/o OGL)~\cite{ezvsl}$_{\text{ECCV}22}$ & \ding{51} &32.66 & 36.72 \\
    & SLAVC(w/o OGL)~\cite{slavc}$_{\text{NeurIPS}22}$ & \ding{51} & 36.50 & - \\
    & \textbf{Ours} & \ding{55} & \textbf{39.11} & \textbf{39.80}\\
    \cline{2-5}
    & \textit{with OGL:}&  &  &  \\
    & EZ-VSL(w/ OGL)~\cite{ezvsl}$_{\text{ECCV}22}$ & \ding{51} &39.57 & 39.60 \\
    & SLAVC(w/o OGL)~\cite{slavc}$_{\text{NeurIPS}22}$ & \ding{51} & 38.87 & - \\
    & \textbf{Ours(w/ OGL)} & \ding{51} & \textbf{42.94} & \textbf{41.54}\\
    \cline{2-5}
    & \textit{with Optical Flow:}&  &  &  \\
    & HearTheFlow~\cite{htf}$_{\text{WACV}23}$& \ding{51} & 39.30 & 40.00 \\
    \bottomrule
    \end{tabular}}
    \caption{\textbf{Comparison results on open-set audio-visual localization experiments trained and tested on the splits of~\cite{chen2021localizing,ezvsl,htf}.}} \label{tab:exp_openset_first}
\end{table}
\begin{table}
    \centering
    \footnotesize
    \scalebox{0.75}{
    \begin{tabular}{l|lccc}
    \toprule
    \textbf{Test Class} & \textbf{Method}   & \textbf{Pre. Vision}  & \textbf{cIoU $\uparrow$}  & \textbf{AUC $\uparrow$} \\ 
    \hline 
    \multirow{2}{*}{Heard 110}  
    & SSSL-TIE~\cite{ssslTransformation}$_{\text{ACM MM}22}$ & \ding{55} & 39.00 & 40.30 \\
    & \textbf{Ours} & \ding{55} &\textbf{41.20} & \textbf{41.00} \\
    \hline \hline

    \multirow{2}{*}{Unheard 110} 
     & SSSL-TIE~\cite{ssslTransformation}$_{\text{ACM MM}22}$& \ding{55} & 36.50 & 38.60 \\
    & \textbf{Ours} & \ding{55} &\textbf{36.90} & 38.59 \\
    \bottomrule
    \end{tabular}}
    \caption{\textbf{Comparison results on open set audio-visual localization experiments trained and tested on the splits of \cite{ssslTransformation}.}} \label{tab:exp_openset_second}
\vspace{-2mm}
\end{table}

We also discuss the methods employed by previous studies, such as SSPL~\cite{song2022sspl} which utilizes a sub-module called PCM to reduce the impact of background noise, HTF~\cite{htf} which utilizes Optical Flow, and EZ-VSL~\cite{ezvsl} which refines its initial audio-visual localization outcomes through object guidance obtained from an ImageNet pretrained visual encoder. Our model, on the other hand, and any of its variations do not require any task-specific modules or operations to achieve the state-of-the-art (SOTA) results. This suggests that using additional semantic and multi-view correspondence, as well as feature alignment, provides more varied and robust supervision for better aligned audio and visual features, as opposed to using task-specific approaches.

The quantitative results presented in~\Tref{tab:quantitative} and~\Tref{tab:quantitative_second} also showcase the performance of previous methods that utilize object guidance to evaluate their final sound source localizations. Our model outperforms all previous methods that employ object guidance on the VGG-SS test set and achieves comparable results on the SoundNet-Flickr test set, even though our model \textit{does not use object guided refinement (OGL)}. Additionally, we acknowledge that the addition of OGL to our audio-visual localization results in improvement on the VGGSS test set, while degrading performance on the SoundNet-Flickr test set. In contrast, prior methods see modest improvements when utilizing OGL. This can be explained by the fact that our model is already accurately localizing the sounding objects, and object guidance can interfere with localization results by introducing visual regions that are not sounding (refer to~\Sref{ssec:qual} for visual results). Unlike prior methods, we do not use OGL in our architecture for the remainder of this paper, unless it is being directly compared with OGL-based methods.

Finally, in comparison to HearTheFlow, which utilizes an additional Optical Flow modality, our method outperforms it on the VGGSS test set, and achieves slightly lower performance on the SoundNet-Flickr test set without utilizing any additional modalities, but instead relying on better audio-visual correspondence and alignment.

\begin{table}[tb!]
\centering
\resizebox{1.0\linewidth}{!}{
\begin{tabular}{c|lcccc}
\toprule
     \textbf{Test Set} & \textbf{Method} & \textbf{Pre. Vision} & \textbf{mIoU $\uparrow$} & \textbf{F-Score $\uparrow$} \\
     \midrule
     \multirow{5}{*}{S4}
     &LVS (w/o OGL)~\cite{chen2021localizing}$_{\text{CVPR}21}$ & \ding{55} &26.9 & 33.6 \\
     &EZ-VSL (w/o OGL)~\cite{ezvsl}$_{\text{ECCV}22}$ & \ding{51}&27.6 & 34.2 \\
     &SLAVC (w/o OGL)~\cite{slavc}$_{\text{NeurIPS}22}$ & \ding{51}&28.1 & 34.6 \\
    \rowcolor{lightgray!25}
    &\myalign{l}{\hspace{-0.6em}\tabnode{\textbf{Ours (w/o OGL)}}} & & & \\
    \rowcolor{lightgray!25}
    &\myalign{l}{\;\;\;\footnotesize $\rotatebox[origin=c]{180}{$\Lsh$}$ NN Search w/ Supervised Pre. Encoders} & \ding{55} & \textbf{29.6} & \textbf{35.9} \\
    \rowcolor{lightgray!25}
    &\myalign{l}{\;\;\;\footnotesize $\rotatebox[origin=c]{180}{$\Lsh$}$ NN Search w/ Self-Supervised Pre. Encoders} & \ding{55} & \underline{29.3} & \underline{35.6} \\
\bottomrule
\end{tabular}
}
{
{\vspace{2mm}
\caption{\textbf{Quantitative results on AVS Bench S4 dataset.} All models are trained on the VGGSound 144K
dataset. }\label{tab:abs_bench}}}
\end{table}

\begin{table}
\tiny
  
  \centering
  \setlength{\tabcolsep}{3pt}
  \begin{tabular}{l c c ccc c ccc }
    \toprule
    \multicolumn{1}{c}{} & \multicolumn{1}{c}{} & \multicolumn{3}{c}{\textbf{A $\rightarrow$ I}} & \multicolumn{3}{c}{\textbf{I $\rightarrow$ A}} \\
    \cmidrule(lr){3-5}\cmidrule(lr){6-8} 
    \textbf{Model}  & \textbf{Pre. Vision}  &\textbf{R@1} & \textbf{R@5} & \textbf{R@10} & \textbf{R@1} & \textbf{R@5} & \textbf{R@10} \\
    \bottomrule
    LVS~\cite{chen2021localizing}$_{\text{CVPR}21}$ & \ding{55} & 3.87 & 12.35 & 20.73 & 4.90 & 14.29 & 21.37\\
    EZ-VSL~\cite{ezvsl}$_{\text{ECCV}22}$ & \ding{51} & 5.01 & 15.73 & 24.81 & 14.2 & 33.51 & 45.18\\
    SSL-TIE~\cite{ssslTransformation}$_{\text{MM}22}$ & \ding{55} & 10.29 & 30.68 & 43.76 & 12.76 & 29.58 & 39.72\\
    SLAVC~\cite{slavc}$_{\text{NeurIPS}22}$ & \ding{51} & 4.77 & 13.08 & 19.10 & 6.12 & 21.16 & 32.12 \\ 
    \rowcolor{lightgray!25}
    \myalign{l}{\hspace{-0.6em}\tabnode{\textbf{Ours}}} & & & & & & & \\
    \rowcolor{lightgray!25}
    \myalign{l}{\;\tiny $\rotatebox[origin=c]{180}{$\Lsh$}$ NN Search w/ Supervised Pre. Encoders} & \ding{55} & \textbf{16.47} & \underline{36.99} & \underline{49.00} & \textbf{20.09} & \textbf{42.38} & \textbf{53.66} \\
    \rowcolor{lightgray!25}
    \myalign{l}{\;\tiny $\rotatebox[origin=c]{180}{$\Lsh$}$ NN Search w/ Self-Supervised Pre. Encoders} & \ding{55} & \underline{14.31} & \textbf{37.81} & \textbf{49.17} & \underline{18.00} & \underline{38.39} & \underline{49.02} \\
    \bottomrule
  \end{tabular}
  \caption{\textbf{Summary of retrieval recall scores for all models.} All of the models are trained on VGGSound 144K data and retrieval is performed on entire VGG-SS dataset, containing \app 5K samples.}\label{tab:retrieval_results}
  \vspace{-4mm}
\end{table}

\newpara{Open Set Audio-Visual Localization.} The study by Chen et al.~\cite{chen2021localizing} evaluates the generalization ability of sound source localization methods in an open set scenario. This involves testing the models on categories that are both present in the training data (heard) and categories that are not present (unheard). To accomplish this, 110 randomly selected categories from the VGGSound dataset are used for training, while another disjoint set of 110 categories are reserved for evaluation to ensure the model have never seen or heard them before. It should be noted that not all previous works use the same train/test splits. While some works, including~\cite{chen2021localizing,ezvsl,slavc,marginnce}, share the same splits,\cite{ssslTransformation} uses a different split. Therefore, to ensure a fair comparison, we conduct experiments on both splits, evaluating on test samples from both heard and unheard categories. The results are shown in~\Tref{tab:exp_openset_first} and~\Tref{tab:exp_openset_second}. Our model outperforms existing approaches on both categories, regardless of the train/test splits used. Specifically, in~\Tref{tab:exp_openset_first}, our model (w/o OGL) even surpasses the other models that use OGL. Previous approaches draw different conclusions from these open set experiments. While some conclude that their models have strong generalization ability because their performance in unheard categories is higher than heard categories~\cite{ezvsl,slavc,marginnce}, the other works that cannot achieve the same trend discuss that this is expected since their models are dealing with unseen categories~\cite{ssslTransformation}. However, our results show that these conclusions are highly dependent on the chosen train/test splits. Our model performs better than existing works in both splits, but there is no uniform trend in between two splits. While our method performs better on unheard categories in the splits of \cite{chen2021localizing,ezvsl,slavc,marginnce}, it performs worse on unheard categories in the split of \cite{ssslTransformation}. Therefore, we conclude that the observed trends are highly dependent on the randomly selected train/test splits.

\newpara{AVSBench~\cite{zhou2022avs}.} To demonstrate the precise sound localization ability of our model, we conduct experiments on the AVSBench S4 dataset. The dataset's objective is to detect audio-visual correspondence and correlation at the pixel level. To make a fair comparison, we use some of the self-supervised sound source localization methods mentioned earlier. All models are trained on VGGSound-144K and directly assessed on the AVSBench S4 dataset without any further fine-tuning (zero-shot setting). Our results, which are presented in~\Tref{tab:abs_bench}, indicate that our method achieves the highest performance, as in the previous experiments.

\begin{table}
    \centering
    \resizebox{1.0\linewidth}{!}{
    \begin{tabular}{lcccccccc}
    \toprule
    &\multicolumn{1}{c}{}& \multicolumn{3}{c}{\textbf{Extended Flickr-SoundNet}} & \multicolumn{3}{c}{\textbf{Extended VGG-SS}} \\
    \textbf{Method} & \textbf{Pre. Vision} &  \textbf{AP $\uparrow$} & \textbf{max-F1 $\uparrow$} & \textbf{LocAcc $\uparrow$} & \textbf{AP $\uparrow$} & \textbf{max-F1 $\uparrow$} & \textbf{LocAcc $\uparrow$} \\ \midrule
    CoarseToFine~\cite{qian2020multiple}$_{\text{ECCV}20}$  & \ding{51} & 0.00 & 38.20 & 47.20 & 0.00 & 19.80 & 21.93  \\
    LVS~\cite{chen2021localizing}$_{\text{CVPR}21}$& \ding{55} & 9.80 & 17.90 & 19.60 & 5.15 & 9.90 & 10.43 \\
    Attention10k~\cite{senocak2018learning}$_{\text{CVPR}18}$&\ding{51} & 15.98 & 24.00 & 34.16 & 6.70 & 13.10 & 14.04 \\
    DMC~\cite{hu2019deep}$_{\text{CVPR}19}$& \ding{51} & 25.56 & 41.80 & 52.80 & 11.53 & 20.30 & 22.63 \\
    DSOL~\cite{hu2020discriminative}$_{\text{NeurIPS}20}$& \ding{51} & 38.32 &  49.40 & 72.91 & 16.84 & 25.60 & 26.87 \\
    OGL~\cite{ezvsl}$_{\text{ECCV}22}$& - & 40.20 & 55.70 & 77.20 & 18.73 & 30.90 & 36.58 \\
    EZ-VSL (w/o OGL)~\cite{ezvsl}$_{\text{ECCV}22}$& \ding{51} & 46.30 & 54.60 & 66.40  & 24.55 & 30.90 & 31.58 \\
    SLAVC (w/o OGL)~\cite{slavc}$_{\text{NeurIPS}22}$& \ding{51}  & 51.63 & 59.10 & \textbf{83.60} & 32.95 & \underline{40.00} & 37.79 \\
    \rowcolor{lightgray!25}
    \myalign{l}{\hspace{-0.6em}\tabnode{\textbf{Ours}}} & & & & & & & \\
    \rowcolor{lightgray!25}
    \myalign{l}{\;\footnotesize $\rotatebox[origin=c]{180}{$\Lsh$}$ NN Search w/ Supervised Pre. Encoders} & \ding{55} & \textbf{64.43} & \textbf{66.90} & \underline{79.60} & \textbf{34.73} & \textbf{40.70} & \textbf{39.94} \\
    \rowcolor{lightgray!25}
    \myalign{l}{\;\footnotesize $\rotatebox[origin=c]{180}{$\Lsh$}$ NN Search w/ Self-Supervised Pre. Encoders} & \ding{55} & \underline{62.67} & \underline{66.10} & 79.20 & \underline{33.09} & \underline{40.00} & \underline{39.20} \\
    \bottomrule
    \end{tabular}}
    {
    \caption{\textbf{Quantitative results on the Extended VGG-SS and Extended SoundNet-Flickr sets}. All models are trained with 144K samples from VGG-Sound. The results of the prior approaches are obtained from~\cite{slavc}.
    }
    \label{tab:extended}}
    \vspace{-4mm}
\end{table}

\newpara{Retrieval.} 
We evaluate sound localization models on the VGG-SS dataset for cross-modal retrieval.
As shown in~\Tref{tab:retrieval_results}, our method clearly outperforms other state-of-the-art methods. 
One interesting observation is that EZ-VSL~\cite{ezvsl} notably performs better than SLAVC~\cite{slavc} on cross-modal retrieval, while SLAVC performs better on sound source localization in~\Tref{tab:quantitative}.
This shows that with the current benchmark evaluations, better sound localization performance does not guarantee better audio-visual semantic understanding, thereby we need to additionally evaluate sound source localization methods on cross-modal understanding tasks.
Another observation is that the performance gap between our method and the strongest competitor SSL-TIE~\cite{ssslTransformation} is notably larger on cross-modal  retrieval than sound source localization.
This is due to the strong cross-modal feature alignment of our method that is overlooked in the sound source localization benchmarks.

\newpara{Extended Flickr and VGG-SS datasets.} The prior study~\cite{slavc} points out that the current sound source localization benchmarks overlook false positive detection. It is because the evaluation samples always contain at least a sounding object in a scene; thus cannot capture false positive outputs, \eg, silent objects or off-screen sounds.
To analyze false positive detection, Mo and Morgado~\cite{slavc} extended the benchmarks with non-audible, non-visible, and mismatched audio-visual samples.
The expectation is that a sound source localization model should not localize any objects when audio-visual semantics do not match.

The experiment with the extended datasets in~\Tref{tab:extended} shows that our method performs favorably against state-of-the-art competitors. Our method performs better than the competing methods in false positive detection measured by $\mathbf{AP}$ and $\mathbf{max}$-$\mathbf{F1}$, while SLAVC~\cite{slavc} achieves better localization performance on Extended Flickr-SoundNet.
As both false positive detection and cross-modal retrieval require cross-modal interaction, our method shows strong performance on both tasks.

\begin{table}[t!]
\centering
\resizebox{1.0\linewidth}{!}{
    \begin{tabular}{cccc|cc}
    \toprule
        & \textbf{Semantic}  & \textbf{Multi-View} & \textbf{Feature Alignment} & \textbf{cIoU $\uparrow$} & \textbf{AUC $\uparrow$} \\
         \midrule
         \rowcolor{lightgray!25}
          (A)&\ding{51}  & \ding{51}  & \ding{51}  & \textbf{39.94}  & \textbf{40.02} \\
          (B)&\ding{51}  & \ding{51}  & \ding{55}  & 39.10  & 39.44 \\
          (C)&\ding{51}  & \ding{55}  & \ding{51}  & 38.75  & 39.34 \\
          (D)&\ding{51}  & \ding{55}  & \ding{55}  & 38.24  & 38.90\\
          (E)&\ding{55}  & \ding{51}  & \ding{51}  & 38.30  & 39.38 \\
          (F)&\ding{55}  & \ding{51}  & \ding{55}  & 37.72  & 39.19 \\
          (G)&\ding{55}  & \ding{55}  & \ding{51}  & 34.93  & 37.94\\
          (H)&\ding{55}  & \ding{55}  & \ding{55}  & 34.22  & 37.67\\
         \bottomrule
    \end{tabular}
}
{
{
\caption{\textbf{Ablation studies on our proposed method to see the impact of each main component.}}\label{tab:main_ablation}}}
\vspace{-2mm}
\end{table}

\subsection{Ablation Results} \label{ssec:ablation} 
We conduct a series of experiments in order to verify our design choices and make further analysis. To save computational time and resources, we primarily perform ablation studies by training our model on VGGSound-144K with NN Search w/ Supervised Pre. Encoders setup and evaluating it on VGG-SS. Results are in~\Tref{tab:main_ablation}.

\begin{figure*}[tp]
    \centering
    \includegraphics[width=\linewidth]{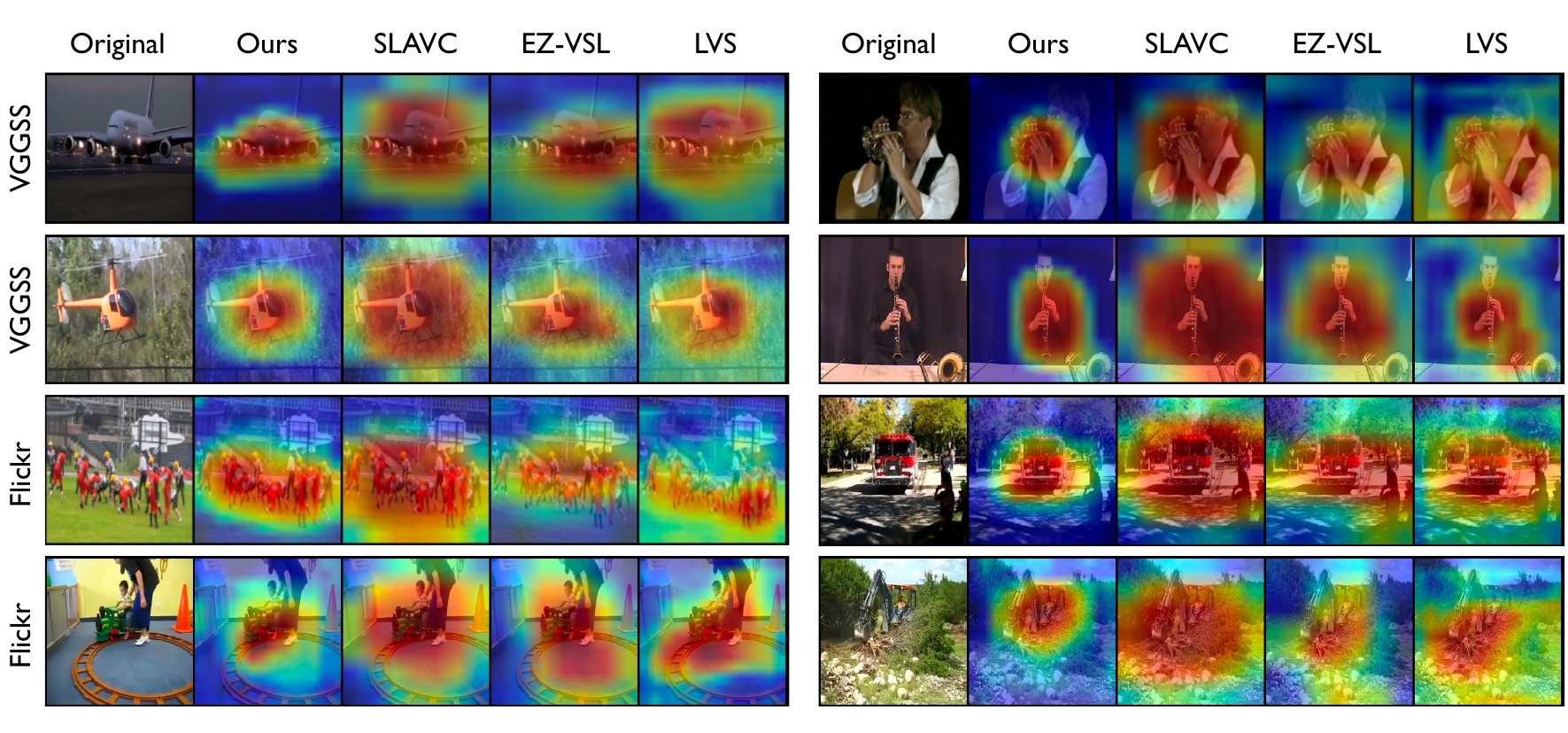}
    \caption{{\bf 
    Sound Localization Results on VGG-SS (top) and SoundNet-Flickr (bottom).
    } }
    \label{fig:qualitatives}
\vspace{-4mm}
\end{figure*}

\newpara{Impact of Semantic and Multi-View Invariance.} In order to understand the impact of each type of invariance (consistency), we analyze the performance of our model with different type of invariance methodologies in~\Tref{tab:main_ablation}. As the results of (C \emph{vs.} E) and (D \emph{vs.} F) reveal, using semantically similar samples (semantic invariance) produces better performance (+0.45$\%$ and +0.5$\%$ on cIoU respectively) compared to augmented multi-view invariance. Moreover, as the results of (A \emph{vs.} C) and (A \emph{vs.} E) depict, the combination of these two different types of invariance complement each other and and further enhances the model's performance. Using pair combination of these two different types of consistency elements provides additional supervisions, invariance and alignments, leading to a more robust representation space and improve sound localization performance. 

\newpara{Impact of Feature Alignment.} We perform controlled experiments to verify the effect of the feature alignment strategy, and the results are presented in~\Tref{tab:main_ablation}. Comparing the performance of the proposed model with and without feature alignment, (A \emph{vs.} B), highlights the importance of this strategy to boost the performance. Further, examining the results of experiments (C \emph{vs.} D) and (E \emph{vs.} F) reveals that feature alignment provides additional gains irrespective of the consistency types. These findings indicate that global feature-based alignment helps the optimization of audio-visual correspondence.

\begin{table}[t!]
\centering
\resizebox{0.9\linewidth}{!}{
    \begin{tabular}{c|cccc>{\columncolor{lightgray!25}}c}
    \toprule
       \textbf{\textit{k} in \textit{k}-NN}  & \textbf{10} & \textbf{30} & \textbf{100} & \textbf{500} & \textbf{1000}\\
         \midrule
         cIoU $\uparrow$ & 38.80  & 38.82 &  39.46 & 39.90 & \textbf{39.94} \\
         AUC $\uparrow$ &  39.51 & 39.67  & 39.93 & 40.00 & \textbf{40.02} \\
         \bottomrule
    \end{tabular}
}
{
{
\vspace{1mm}
\caption{\textbf{Varying k in conceptually similar sample selection.}}\label{tab:topk}}}
\vspace{-4mm}
\end{table}

\newpara{Impact of $k$ in conceptually similar sample selection.} Selecting an appropriate $k$ value for sampling nearest neighbors is crucial. If this value is set too high, it may result in noisy samples that could disrupt the learning phase. Conversely, if the value is set too low, only very similar samples to the anchor will be provided and it limits semantic invariance.
Nevertheless, when compared to \Tref{tab:main_ablation} (E), we observe performance gain throughout the range of $k$ used for the ablation study.
~\Tref{tab:topk} shows an ablative evaluation of the effect of $k$ value used to select neighborhood samples. 
The results indicate that an optimal choice is $k$=1000. This choice of $k$ can be explained by the fact that it provides a balance between semantic similarity and sufficient diversity.

\subsection{Qualitative Results} \label{ssec:qual}
In this section, we visualize and compare our sound localization results with the recent prior works on standard benchmarks, namely on VGG-SS and SoundNet-Flickr. 
The visualized samples in~\Fref{fig:qualitatives} show that localized regions of the proposed method are more compact and accurately aligns with the sounding objects than the other methods. For instance, small size musical instrument is localized accurately compared to the recent methods in the top right column.

\begin{figure}[tp]
    \centering
    \includegraphics[width=\linewidth]{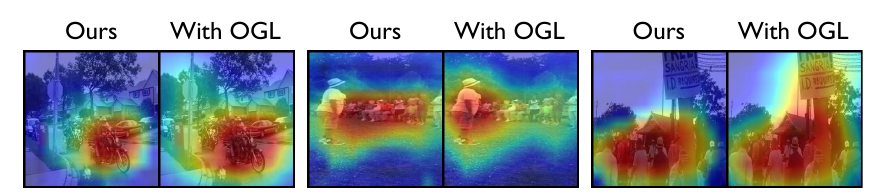}
    \caption{{\bf 
    OGL degrades our sound localization results on SoundNet-Flickr.
    } }
    \label{fig:ogl}
\vspace{-4mm}
\end{figure}

\begin{figure}[tp]
    \centering
    \includegraphics[width=\linewidth]{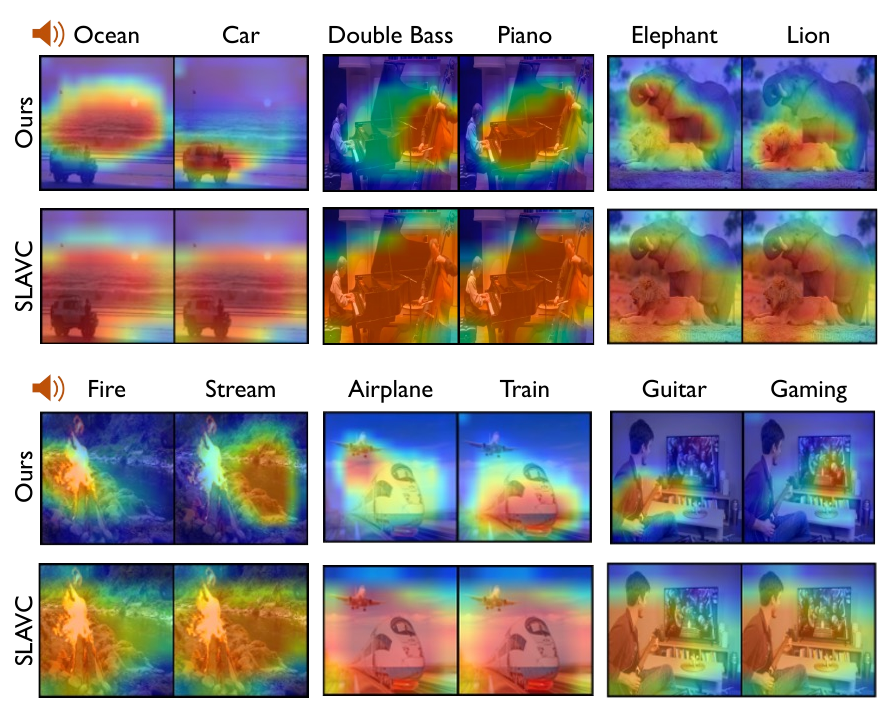}
    \caption{{\bf Interactive Sound Localization of Ours and SLAVC~\cite{slavc}.
    } Our model correctly follows the cross-modal interaction for given different sounds.}
    \label{fig:interactive}
\vspace{-4mm}
\end{figure}

We also compare our localization results with and without object-guided localization (OGL). 
As shown in \Fref{fig:ogl}, OGL deteriorates our sound localization outputs. 
OGL captures objectness in a scene, thereby tending to attend to any distinctive objects regardless of whether it is the sound source or not.
Therefore, OGL can be helpful when localization totally fails because of the objectness bias in the benchmarks, but it is harmful when the localization is accurate which is the case for the examples shown.
This result is consistent with the quantitative result in~\Tref{tab:quantitative_second}, showing that our method with OGL performs worse.

Throughout the paper, we discuss the importance of cross-modal semantic understanding. We demonstrate interactiveness of our method across modalities in~\Fref{fig:interactive}.
Genuine sound source localization should be able to localize objects that are correlated with the sound. To visualize cross-modal interaction, we synthetically pair up the same image with different sounds of objects that are visible in a scene. The examples demonstrate that the proposed method can localize different objects depending on the contexts of sounds, while the competing method can not.

\section{Conclusion}
In this work, we investigate cross-modal semantic understanding that has been overlooked in sound source localization studies.
We observe that higher sound source localization performance on the current benchmark does not necessarily show higher performance in cross-modal retrieval, despite its causal relevance in reality.
To enforce strong understanding of audio-visual semantic matching while maintaining localization capability, we propose semantic alignment with multi-views of audio-visual pairs in a simple yet effective way.
The ablation study shows that strong semantic alignment is achieved when both semantic alignment loss and enriched positive pairs are used.
We extensively evaluate our method on sound source localization benchmarks including cross-dataset and open-set settings. Moreover, our analyses on cross-modal retrieval and false positive detection verify that the proposed method has strong capability in cross-modal interaction.
Our study suggests that sound localization methods should be evaluated not only on localization benchmarks but also on cross-modal understanding tasks.
\section{Acknowledgment}
This work was supported by the National Research Foundation of Korea (NRF) grant funded by the Korea government (MSIT) (No. RS-2023-00212845, Multi-modal Speech Processing for Human-Computer Interaction). 
H.~Pfister and J.~Kim were partially supported by NIH grant R01HD104969.
T.-H. Oh was partially supported by IITP grant funded by the Korea government (MSIT) (No. 2021-0-02068, Artificial Intelligence Innovation Hub; No. 2022-0-00290, Visual Intelligence for Space-Time Understanding and Generation based on Multi-layered Visual Common Sense).

{
\small
\bibliographystyle{template/ieee_fullname}
\bibliography{egbib}
}

\end{document}